\documentclass[letterpaper, 10 pt, conference]{ieeeconf}  

\IEEEoverridecommandlockouts                              

\pdfoutput=1
\usepackage{amsmath}
\usepackage{caption}
\usepackage{subcaption}
\usepackage{graphicx}
\usepackage{hyperref}
\usepackage{cleveref}
\usepackage{url}
\usepackage{xcolor}
\usepackage{tikz}
\usepackage{multirow}
\usepackage{makecell}

\usepackage{mathptmx}
\DeclareCaptionFont{times8}{\fontfamily{ptm}\fontsize{8}{10}\selectfont}
\usepackage[T1]{fontenc}
\title{\LARGE \bf
	An Integrating Comprehensive Trajectory Prediction with Risk Potential Field Method for Autonomous Driving
}
\author{Kailu Wu, Xing Liu, Feiyu Bian, Yizhai Zhang, and Panfeng Huang
	\thanks{*This work was supported in part by the National Key R\&D Program of China under Grant 2022ZD0117900, and the National Natural Science Foundation of China under Grant 62103334, 92370123 and 62273280.}%
	\thanks{Kailu Wu, Xing Liu, Feiyu Bian, Yizhai Zhang, and Panfeng Huang(Corresponding author) are with the Research Center for Intelligent Robotics, School of Astronautics, Northwestern Polytechnical University, and National Key Laboratory of Aerospace Flight Dynamics, Northwestern Polytechnical University, Xi’an, China, 710072 e-mail: pfhuang@nwpu.edu.cn.}%
}

\begin{document}
	\maketitle
	\thispagestyle{empty}
	\pagestyle{empty}

	\begin{abstract}
		Due to the uncertainty of traffic participants' intentions, generating safe but not overly cautious behavior in interactive driving scenarios remains a formidable challenge for autonomous driving. In this paper, we address this issue by combining a deep learning-based trajectory prediction model with risk potential field-based motion planning. In order to comprehensively predict the possible future trajectories of other vehicles, we propose a target-region based trajectory prediction model(TRTP) which considers every region a vehicle may arrive in the future. After that, we construct a risk potential field at each future time step based on the prediction results of TRTP, and integrate risk value to the objective function of Model Predictive Contouring Control(MPCC). This enables the uncertainty of other vehicles to be taken into account during the planning process. Balancing between risk and progress along the reference path can achieve both driving safety and efficiency at the same time. We also demonstrate the security and effectiveness performance of our method in the CARLA simulator.
	\end{abstract}

	\section{INTRODUCTION}
	The technology of autonomous vehicle(AV) has made significant advancements in recent years. However, navigating safely and efficiently in interactive scenarios still remains a particularly difficult challenge due to the uncertainty of traffic participants’ intentions. Predicting the future trajectories of traffic participants can assist autonomous vehicles in understanding potential future traffic conditions. Moreover, ensuring driving safety and efficiency requires incorporating predictions of future situations into planning processes. Therefore, trajectory prediction and the combination of prediction and planning are crucial for AVs.
	
	Predicting the future trajectories of vehicles in interactive traffic scenarios is challenging because of the multi-modal nature of drivers' intentions. An important task in trajectory prediction is to make the prediction results as comprehensive as possible to avoid traffic accidents due to failure to consider certain situations. Some methods obtain multi-modal predictions by outputting a fixed number of trajectories \cite{cui2019multimodal}, \cite{deo2022multimodal}, \cite{mercat2020multi}. Nonetheless, employing a fixed number of trajectories does not guarantee comprehensive coverage of all potential modes of other vehicles. Moreover, fixed trajectories lack flexibility and cannot adapt to changing traffic conditions. Recently, some researchers have effectively obtained multi-modal trajectory prediction using the driver's goals, where the goals are represented as target lanes \cite{luo2020probabilistic}, \cite{zhang2021map}. But the target lane can not include all possibilities, as the driver's position within the lane may vary during this prediction horizon due to different driving modes, such as acceleration, deceleration, or stopping. In order to encompass all potential modes in the future, we also propose a goal-based prediction model TRTP. The distinction lies in representing the target as the regions the vehicle aims to reach within T seconds, which enables a more comprehensive consideration.
	
	\begin{figure}[tb]
		\centering
		\begin{subfigure}{\linewidth}
			\centering
			\includegraphics[width=\textwidth]{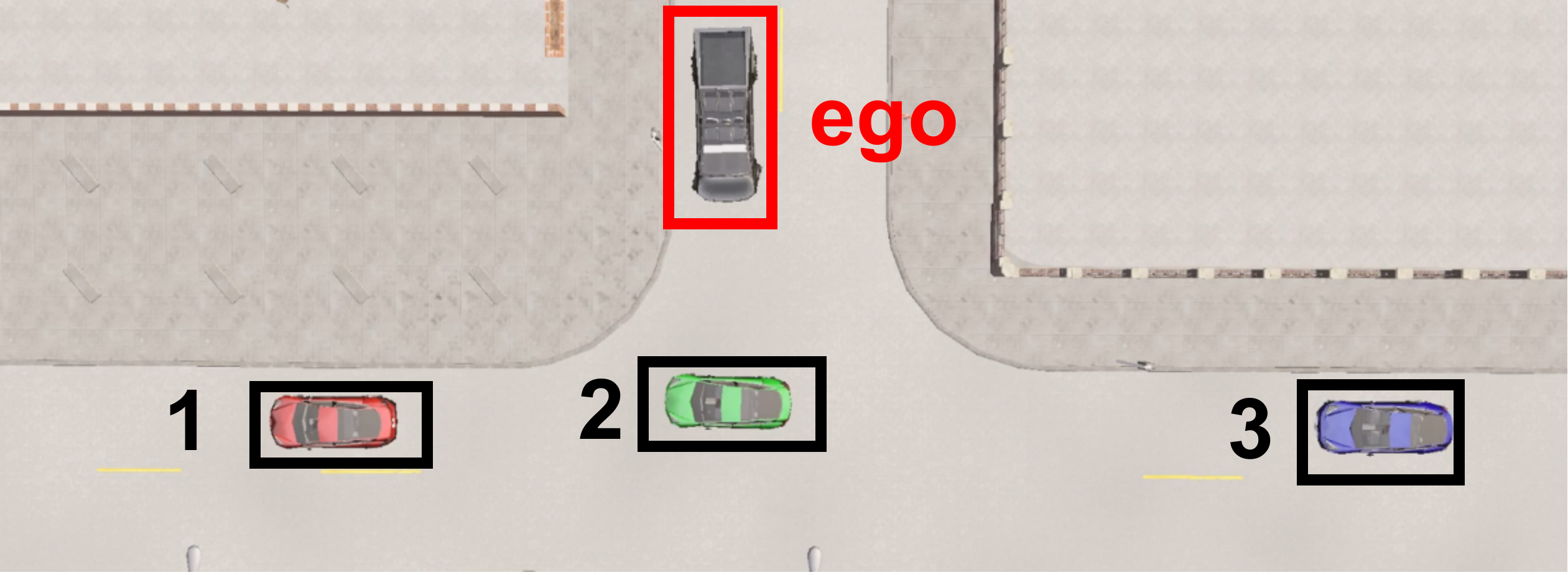}
			\caption{}
			\label{fig:fulla}
		\end{subfigure}
		\begin{subfigure}{\linewidth}
			\centering
			\includegraphics[width=\textwidth]{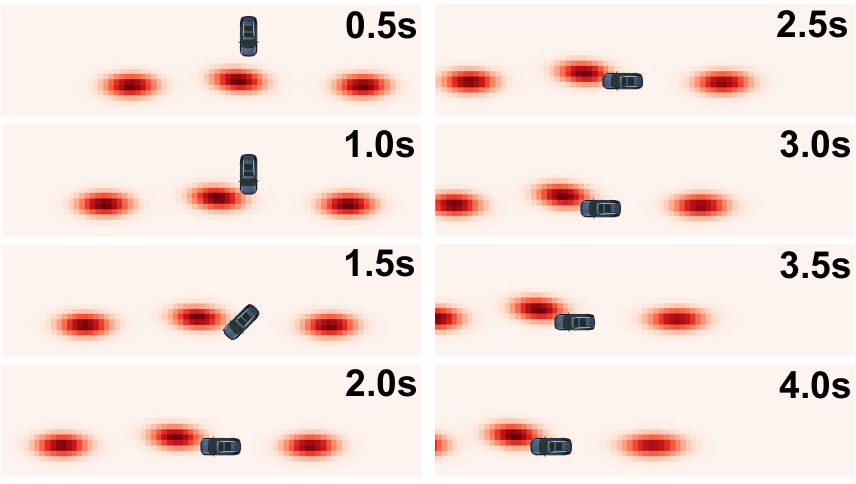}
			\caption{}
			\label{fig:fullb}
		\end{subfigure}
		\caption{Illustration of the risk potential fields and planning results at different time steps in the next 4 seconds. (a)At the current moment, the vehicle is preparing to merge into the traffic flow. (b) In the next 4 seconds, the risk potential field of each time step is established based on the results obtained by TRTP. The car in the risk potential field represents the pose of the ego vehicle at each time step in the future planned by the motion planning module mpcc.}
		\label{fig:full description}
	\end{figure}

	 At present, methods based on numerical optimization are most commonly used for planning in autonomous vehicles, such as function optimization and model prediction methods \cite{badue2021self}. Yet, integrating the prediction results obtained from deep learning-based models with optimization-based planning modules for safe and effective driving remains a challenge. In order to integrate these two types of approaches, we establish risk potential fields for the future based on the predictions of TRTP, and conduct planning within this field. A high-performance controller for autonomous racing is obtained by Model Predictive Contouring Control(MPCC) in \cite{liniger2015optimization}, enabling trajectory tracking along the reference path. Because we aim to ensure safety and efficiency while navigating along the reference path, MPCC is a good choice for considering the future. Therefore, we incorporate the risk values from the risk potential field into the objective function of MPCC to achieve a balance between these two objectives.
	
	Our contributions are summarized as follows:
	\begin{itemize}
		\item We propose a trajectory prediction model TRTP based on target regions that a vehicle may reach in the future, which can get comprehensive trajectory predictions.
		\item We construct risk potential fields based on the results of TRTP and incorporate the risk value into the MPCC objective function to achieve non-conservative risk-aware motion planning.
	\end{itemize}
	
	\section{RELATED WORKS}
	\subsection{Trajectory prediction}
	Trajectory prediction is very important for ensuring the safety of AV. Physics-based motion models are used to predict trajectories such as dynamic models and kinematic models. But they are limited to short-term(less than a second) prediction, which is not enough for safe driving \cite{lefevre2014survey}. Since deep learning-based methods can capture the driver's hidden state and interaction information, an increasing number of researchers are utilizing them to accomplish prediction tasks, resulting in remarkable achievements. As mentioned above, obtaining multi-modal predictions of the target vehicle is crucial for the safety of AVs. So we will give a more detailed discussion below.
	
	In order to get multiple possible trajectories that the vehicle may travel in the future, models are designed as a regression problem \cite{cui2019multimodal, liang2020learning}. Some methods also use the ideas of classification to predict trajectories. MultiPath \cite{chai2019multipath} performs classification over the anchors and regresses offsets from anchor waypoints along with uncertainties. CoverNet \cite{phan2020covernet} classifies predicted trajectories on a trajectory set. Goal-net \cite{zhang2021map} and CXX \cite{luo2020probabilistic} classify target paths represented by target lanes' centerlines. There are also some sample-based methods to approximate the distribution over future behavior. DESIRE \cite{lee2017desire} combines a deep generative model which introduces a latent variable to account for the ambiguity of the future with a past observations encoder to generate multiple prediction hypotheses. MATF \cite{zhao2019multi} use Generative Adversarial Networks(GANS) to capture the distribution over predicted trajectories with a set of samples. The drawback of the sample-based models is that they are not easy to deploy because a substantial number of samples are needed to comprehensively capture the distribution of future behaviors.
	
	High Definition Maps (HD maps) provide detailed geometric information about lanes, crosswalks, stop signs, and more for trajectory prediction. In order to effectively extract the information in the HD map, VectorNet \cite{gao2020vectornet} represents agent trajectories and map features as sequence of vectors, and then processed by  graph neural networks. The resulting feature information is passed to a fully connected graph to model higher-order interactions. LaneGCN \cite{liang2020learning} constructs a lane graph from HD maps to extract lane features and uses  convolution to extract actor features. Then the features  from both actors and the map are fused via a FusionNet which avoids information loss.
	
	\subsection{Motion planning based on prediction}
	Having obtained predictions of other traffic participants, the objective of motion planning for autonomous driving is to generate a trajectory that ensures safety, efficiency, and successful arrival at the destination. Some methods make autonomous vehicles avoid the forward reachable set \cite{kousik2017safe} or all possible trajectories of other agents \cite{chen2021reactive}. This type of approach could lead to vehicles exhibiting overly cautious behavior, potentially resulting in traffic congestion and increased likelihood of accidents. Contingency planning \cite{hardy2013contingency} predict other agents' state distributions and treats planning problems as constrained numerical optimization problems. MARC \cite{li2023marc} and branch Model Predictive Control(MPC) \cite{chen2022interactive} plan trajectory based on the scenario tree which is constructed from the semantic-level intentions of other agents. \cite{pan2020safe} incorporates short-term and long-term predictions of target vehicles to ensure safety and efficiency in planning. \cite{wang2023interaction} predicts the surrounding vehicle's trajectory by learning its level of cooperation online and integrates the predictions with the MPC algorithm to ensure safety. 
 	
	\section{TRAJECTORY PREDICTION MODEL}
\begin{figure*}[tb]
		\centering
		\begin{subfigure}{0.28\textwidth}
			\centering
			\includegraphics[width=\textwidth]{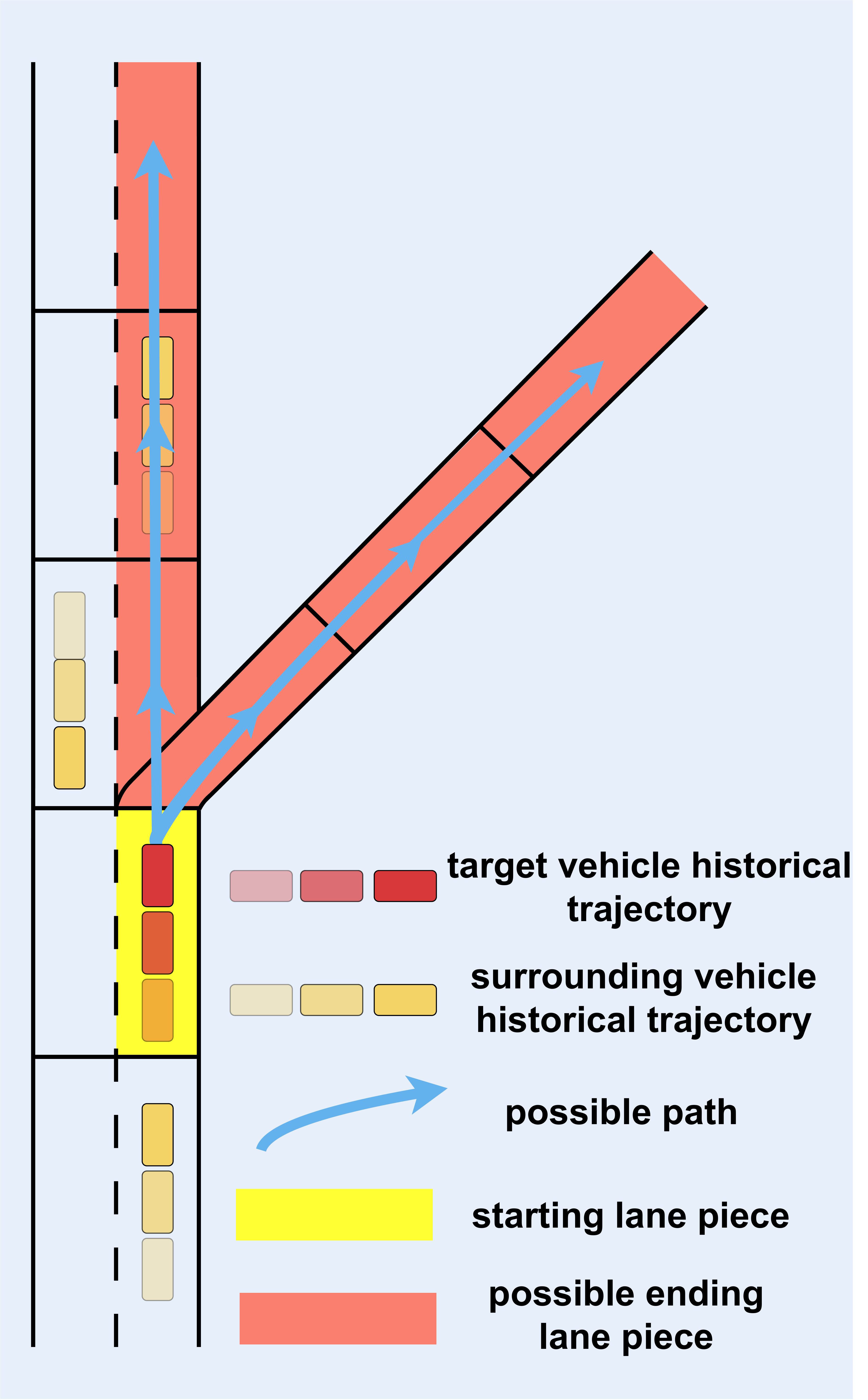}
			\caption{}
			\label{fig:pre_model1}
		\end{subfigure}
		\begin{subfigure}{0.66\textwidth}
			\centering
			\includegraphics[width=\textwidth]{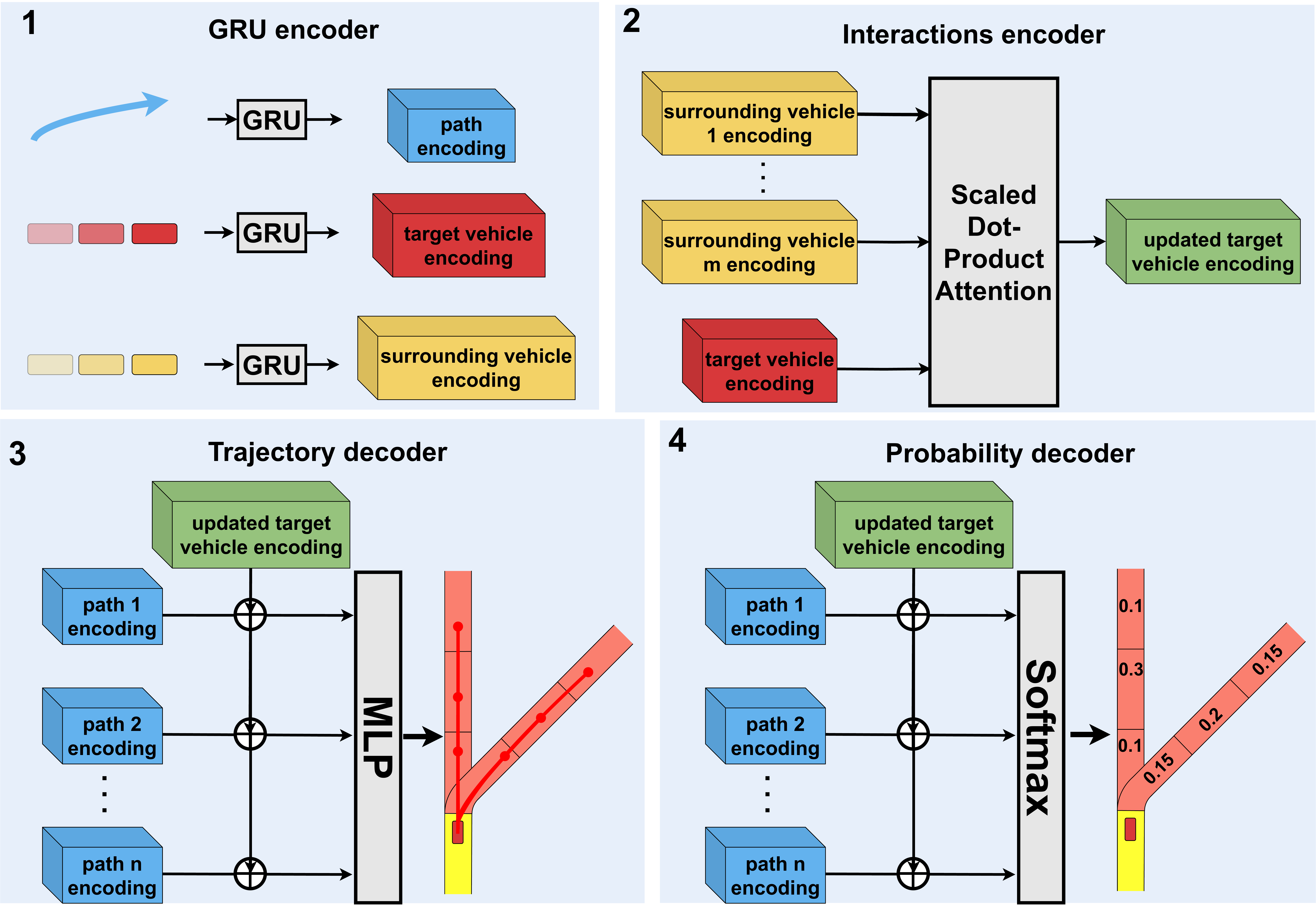}
			\caption{}
			\label{fig:pre_model2}
		\end{subfigure}
		\caption{Illustration of TRTP. (a) shows that lanes are split into small pieces of equal length. Currently, the target vehicle is within the yellow lane piece, and after T seconds, it may be located in the red ones. Possible target paths are generated based on the topological connection relationships between lane pieces and are represented by blue arrows. (b) is the architecture of TRTP. TRTP consists of four modules. GRU encoder encodes the possible paths and historical trajectories of vehicles. Interactions encoder uses Scaled Dot-Product Attention to represent interactions between vehicles. The trajectory decoder and probability decoder decode a trajectory and corresponding probability for each path respectively.}
		\label{fig:pre_model}
	\end{figure*}

	The prediction of potential future trajectories for a vehicle is influenced by a myriad of factors, including its historical trajectory, lane constraints, inter-vehicle interactions, and other pertinent elements. Understanding and modeling these factors are crucial for accurate and reliable predictions in dynamic traffic scenarios. As shown in \autoref{fig:pre_model}, TRTP takes these influencing factors into account.
	
	\subsection{Problem Statement}
	We represent the historical trajectory of vehicle i as a vector \(H_{i}=[X_{t_{0}-s}^{i},\cdots,X_{t_{0}-1}^{i},X_{t_{0}}^{i}]\) where \(t_{0}\) and \(s\) represent the current time step and historical duration respectively. \(X_{t}^{i}=[x_{t}^{i},y_{t}^{i},\varphi_{t}^{i},v_{t}^{i},a_{t}^{i},w_{t}^{i}]\) is the state variable, where \(x_{t}^{i}\), \(y_{t}^{i}\), \(\varphi_{t}^{i}\) are the location and yaw, \(v_{t}^{i}\), \(a_{t}^{i}\), \(w_{t}^{i}\) are the speed, acceleration and yaw rate of the vehicle \(i\) at time \(t\). As shown in \autoref{fig:pre_model}, we divide lanes around the target vehicle into small pieces of equal length. Each lane piece which is represented as \(N_{j}=\left \{ C_{0}, C_{1},\cdots,C_{N}\right \}\) and  \(C_{N}=\left [ x^{N},y^{N},\varphi ^{N}\right ] \) is the coordinate and yaw of a discrete point of this lane piece's centerline. At time \(t_{0}\) the target vehicle is located in certain starting lane pieces \(N_{s}=\left \{ N_{s_{0},}\cdots,N_{s_{m} }\right \} \). The possible ending lane pieces that the target vehicle may reach after T seconds is  \(N_{e}=\left \{ N_{e_{0},}\cdots,N_{e_{n} }\right \} \). According to the topological connection relationship between the lane pieces, we obtain a set of possible paths for the target vehicle to travel in the future which is denoted by \(P=\left \{P_{1}, P_{2},\cdots ,P_{k}\right \} \). One path in the set is \(P_{k}=\left[ N_{s_{i}},N_{e_{j_{0}}},\cdots,N_{e_{j_{l} } } \right ] \), where \(N_{s_{i}}\) is the starting lane piece, \(N_{e_{l}}\) is the ending lane piece and the pieces in path \(P_{k}\) are connected to each other. Our aim is to get the set of predicted trajectories \(\tau=\left \{ \tau _{0},\tau _{1},\cdots,\tau _{k}\right \} \) and the corresponding probabilities \(p=\left \{ p_{0},p_{1},\cdots,p_{k} \right \} \) for the target vehicle. \(\tau _{k}=\left [X_{k}^{1},X_{k}^{2},\cdots,X_{k}^{T}\right ] \) is one of the predicted trajectories where \(X_{k}^{t}=\left [x_{k}^{t},y_{k}^{t},\varphi_{k}^{t}\right] \) denotes the predicted pose at time step \(t\).
	\subsection{Encoder}
	We use the Gated Recurrent Unit(GRU) to encode the historical trajectory \(H_{t}\) of the target vehicle and the historical trajectories \(H_{i}\) of its surrounding vehicles. Since the interaction between vehicles has a significant impact on trajectory prediction, we use the encoding of surrounding vehicles to update the target vehicle encoding \(E_{updated}^{t} \) through the Scaled Dot-Product Attention \cite{vaswani2017attention}. We also encode every path \(P_{k}\) in the path set \(P\) to get \(E_{path}^{k}\) by GRU.
	\subsection{Decoder}
	Getting the updated encoding of the target vehicle \(E_{updated}^{t} \) and the encoding of every path \(E_{path}^{k}\), we use Multilayer Perceptron(MLP) to decode the future trajectories that each path in the path set will decode one trajectory. In this way, all lane pieces that the target vehicle may reach in the next T seconds are considered. In order to obtain the probability corresponding to each trajectory, we put the target vehicle encoding \(E_{updated}^{t} \) and the path encoding \(E_{path}^{k}\) into the Scaled Dot-Product Attention module to decode the probabilities.
	\begin{equation}
		\tau _{k}= MLP(concat(E_{updated}^{t},E_{path}^{k}))
	\end{equation}
	\begin{equation}
		p_{k}= Attention(E_{updated}^{t},E_{path}^{k})
	\end{equation}
	\subsection{Loss Function}
	In the path set P, the path \(P^{*} \) has the same ending lane piece as the ground truth trajectory \(\tau _{gt} =\left \{ X_{gt}^{0},\cdots,X_{gt}^{T}\right \} \) of the target vehicle in the future T seconds. We represent the predicted trajectory decoded by \(P^{*} \) as \(\tau ^{*} \) and its associated probability as \(p^{*} \). The loss function for the trajectory decoder is the average displacement error between  \(\tau ^{*} \) and \(\tau _{gt} \):
	\begin{equation}
		L_{traj}=\sum_{i=0}^{T}\left \| X_{*}^{i}-X_{gt}^{i}\right \|_{2}
	\end{equation}
	And the loss function for the probabilities decoder is 
	\begin{equation}
		L_{pro}=-log\left (p^{*}   \right )
	\end{equation}
	The final loss function for our model is
	\begin{equation}
		L= \alpha L_{traj}+ \beta L_{pro}
	\end{equation}
	\section{Motion planning}\label{sec:Motion planning}
	Obtaining the trajectory predictions of vehicles around ego vehicle, we can plan a safe trajectory taking into account the future risks in advance. We generate a risk potential field at each future time step based on the results of TRTP and incorporate it into the MPCC objective function to get a risk-aware motion planning for ego vehicle.
	
	\subsection{Vehicle Model}
\begin{figure}[tb]
		\centering
		\includegraphics[width=0.6\linewidth]{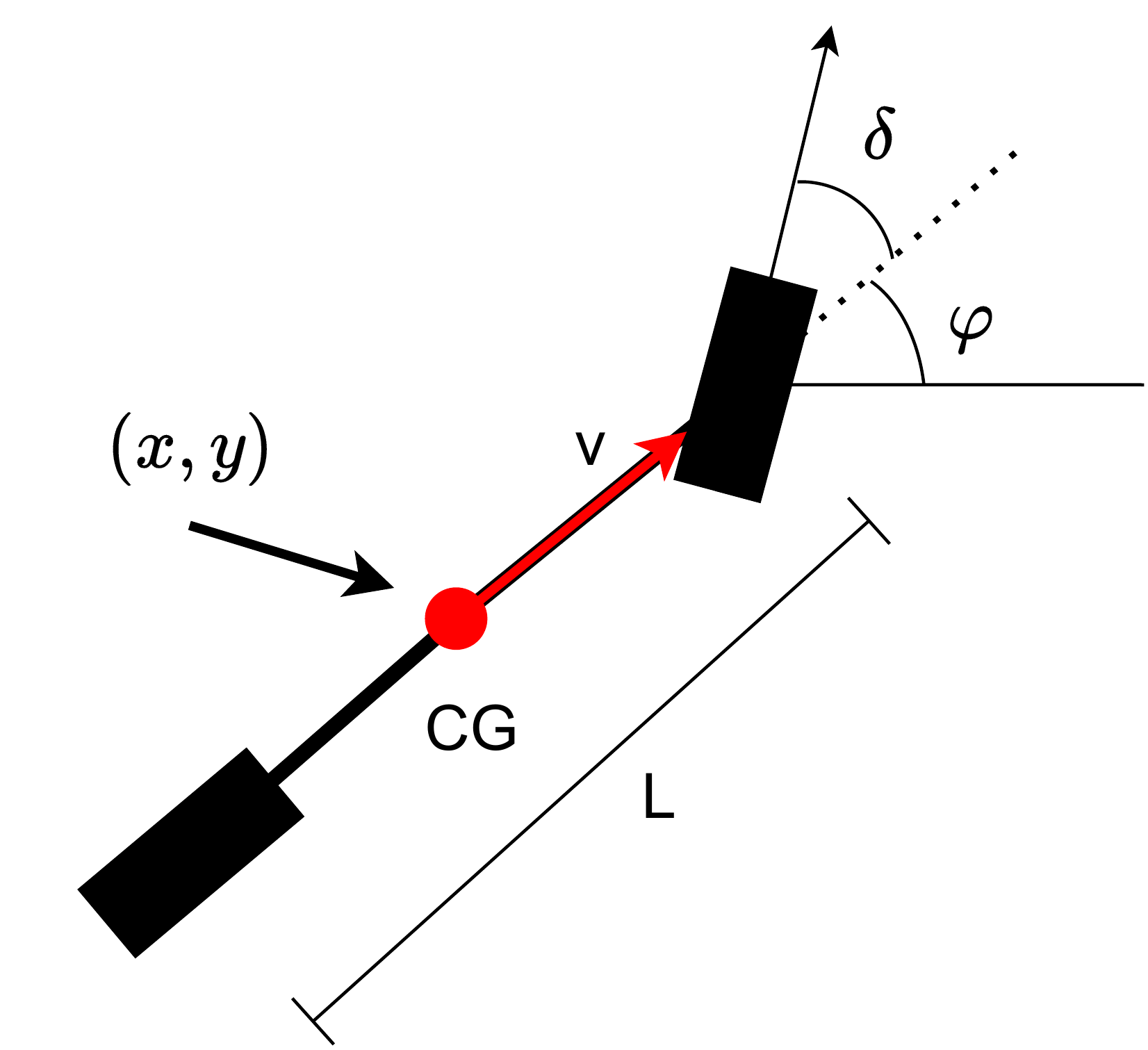}
		\caption{Kinematic bicycle model.}
		\label{fig:vehicle_model}
	\end{figure}

	We model the vehicle dynamics using a rear-wheel drive kinematic bicycle model as shown in \autoref{fig:vehicle_model}. The state space description of the model is derived around the center of gravity(CG), where \(L\) is the distance between the two wheel axes and \(\delta\) is the steering angle for the front wheel. The state vector \(\mathbf{x}=[x,y,\varphi ,\delta ,v]^{T} \) consists of the vehicle's position \(x\), \(y\), heading angle \(\varphi\), steering angle \(\delta\) and longitudinal speed \(v\). The control input \(\mathbf{u}=[\dot{\delta} ,a]^{T}\) is composed of steering velocity \(\dot{\delta}\) and longitudinal acceleration \(a\).
	
	\begin{equation}
		\begin{bmatrix}
			\dot{x}\\ 
			\dot{y}\\
			\dot{\varphi }\\ 
			\dot{\delta }\\ 
			\dot{v}\\ 
		\end{bmatrix}=\begin{bmatrix}
			vcos\varphi\\ 
			vsin\varphi\\ 
			\frac{v}{L} tan\delta\\ 
			0\\
			0
		\end{bmatrix}+\begin{bmatrix}
			0&0\\
			0&0\\
			0&0\\
			1&0\\
			0&1
		\end{bmatrix}\begin{bmatrix}
			\dot{\delta }\\ 
			a \label{eq:model}
		\end{bmatrix}
	\end{equation}
	
	\subsection{Risk Potential Field}
	At current time step \(t_{0}\), there are m vehicles that need to be predicted. For vehicle \(i\) we get the predicted trajectories and the corresponding probabilities through the prediction model mentioned above. The predicted global pose of trajectory \(j\) at time \(t\) is \(X_{j}^{t}=\left [x_{j}^{t},y_{j}^{t},\varphi_{j}^{t}\right] \). The potential field caused by vehicle \(i\) at future time step \(t\) is generated by exponential function:
	\begin{equation}\label{eq:rbf point}
		\Theta\left ( x,y \right )  _{i}^{t}=\sum_{j=0}^{k}p_{j}e^{-(\frac{\Delta x_{j}^{2} }{a^{2} }+\frac{\Delta y_{j}^{2} }{b^{2} }) }    
	\end{equation}
	In \eqref{eq:rbf point}, \(a\),\(b\) are shape parameters of the risk potential field and \(p_{j}\) is the probability of the jth predicted trajectory. Moreover, \(\Delta x_{j} \), \(\Delta y_{j} \) are the components of the Euclidean distance between \(X_{j}^{t} \) and \((x,y) \) in the local coordinate system of the target vehicle, calculated using the following equation:
	
	\begin{equation}
		\begin{bmatrix}
			\Delta x_{j} \\
			\Delta y_{j}
		\end{bmatrix}=\begin{bmatrix}
			\cos \varphi &\sin \varphi \\
			-\sin\varphi &\cos \varphi 
		\end{bmatrix}\begin{bmatrix}
			x-x_{j} \\
			y-y_{j} 
		\end{bmatrix}
	\end{equation}
	Therefore, the risk potential field formed by all agents at time \(t\) is:
	\begin{equation}
		\Theta(x,y) ^{t}=\gamma ^{t} \sum_{i=1}^{m}\Theta(x,y) _{i}^{t}  
	\end{equation}
	where \(\gamma\) is a discount factor, which means that the longer the time from the current moment, the smaller the impact of the risk value on the planning. The risk potential fields generated based on TRTP are shown in \autoref{fig:full description}.
	
	\subsection{Model Predictive Contouring Control}
	Given a starting point and an ending point, the A* algorithm can plan a global reference path based on map features. The reference path is parameterised by the arc-length \(\theta\) as \((X^{ref} (\theta ),Y^{ref}(\theta )\). We can get the angle of any point on the reference path by:
	\begin{equation}\label{eq:ref_theta}
		\phi(\theta )=arctan(\frac{\partial Y(\theta )}{\partial X(\theta )} )  
	\end{equation}
	To enable the vehicle to follow the reference path, it is necessary to define the deviation between the vehicle position \((X_{k},Y_{k})\) and the reference path. The contouring error \(e_{k}^{c}\) is defined as the distance between them, whose expression is: 
	\begin{equation}
		\begin{split}
			e_{k}^{c} &= \sin(\Phi(\theta _{r})) (X_{k}^{ref}(\theta _{r})-X_{k}) \\
			&\quad - \cos(\Phi(\theta _{r})) (Y_{k}^{ref}(\theta _{r})-Y_{k})
		\end{split}
	\end{equation}
	\begin{figure}[tb]
		\centering
		\includegraphics[width=\linewidth]{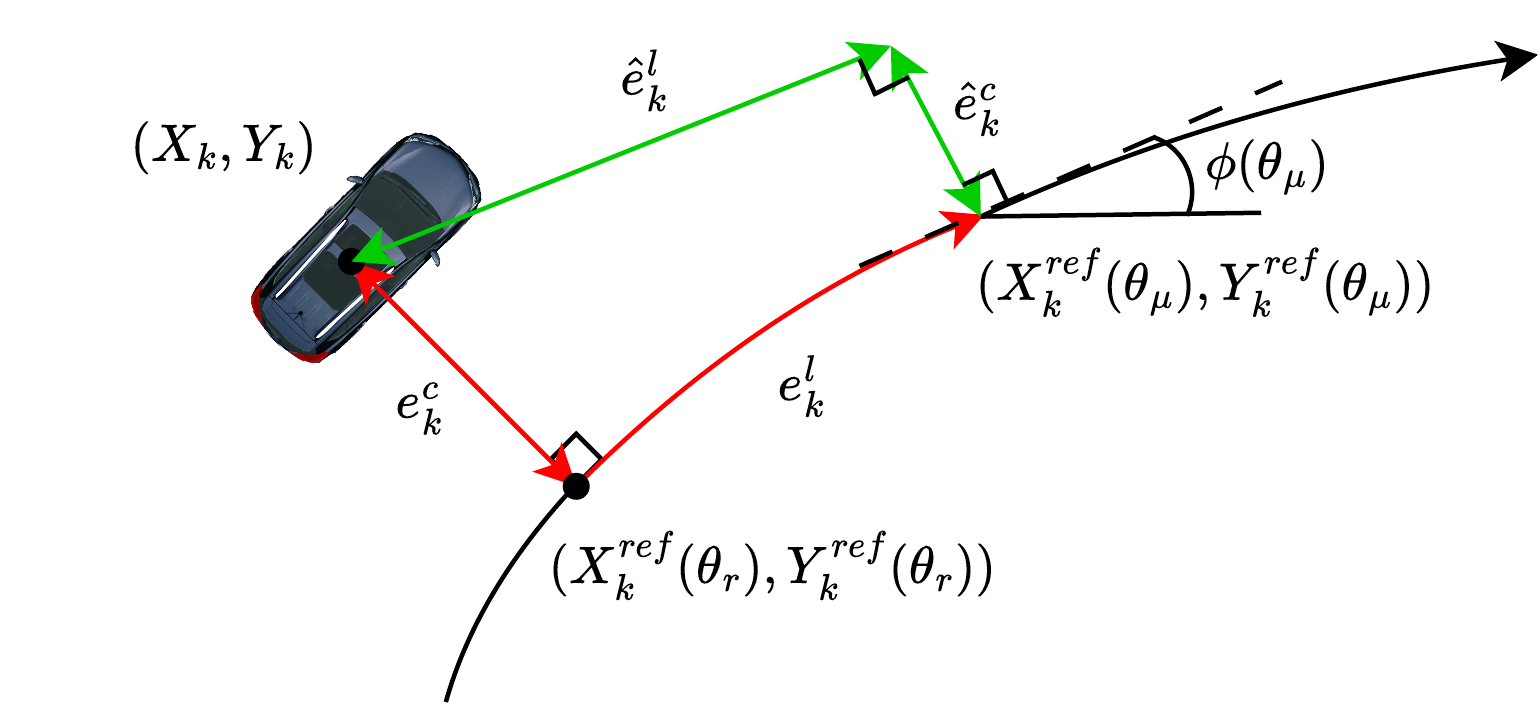}
		\caption{\(\theta _{r}\) is the projection of vehicle position onto the reference path. \(\theta _{\mu}\) is an approximation of \(\theta _{r}\) and the contouring error \(e_{k}^{c}\) is approximated by \(\hat{e}_{k}^{c}\), the lag error \(e_{k}^{l}\) is approximated by \(\hat{e}_{k}^{l}\).}
		\label{fig:mpcc}
	\end{figure}
	As shown in \autoref{fig:mpcc}, \((X_{K}^{ref}(\theta _{r } ), Y_{K}^{ref}(\theta _{r } ))\) is the closet point to \((X_{k},Y_{k})\) on the reference path. In optimization problems, it is not feasible to directly calculate the position of \((X_{k}^{ref}(\theta _{r } ), Y_{k}^{ref}(\theta _{r } ))\) online due to the limitation of calculation speed. Therefore, \((X_{k}^{ref}(\theta _{\mu } ), Y_{k}^{ref}(\theta _{\mu } ))\) is used to approximate it and at the same time the lag error \(e_{k}^{l}\) is introduced to measure the quality of approximation. The contouring error \(e_{k}^{c} \) can then be approximated by:
	\begin{equation}
		\begin{split}
			\hat{e}_{k}^{c} &= \sin(\Phi(\theta _{\mu})) (X_{k}^{ref}(\theta _{\mu})-X_{k}) \\
			&\quad - \cos(\Phi(\theta _{\mu})) (Y_{k}^{ref}(\theta _{\mu})-Y_{k})
		\end{split}
	\end{equation}
	The lag error \(e_{k}^{l} \) can then be approximated as:
	\begin{equation}
		\begin{split}
			\hat{e}_{k}^{l} &= \cos(\Phi(\theta _{\mu}) (X_{k}^{ref}(\theta _{\mu})-X_{k})) \\
			&\quad + \sin(\Phi(\theta _{\mu})) (Y_{k}^{ref}(\theta _{\mu})-Y_{k}))
		\end{split}
	\end{equation}
	At each time step, the approximate parameter updated along the reference path is:
	\begin{equation}
		\theta _{\mu ,k+1} =\theta _{\mu ,k}+ v_{\theta, k } \Delta t
	\end{equation}
	where \(v_{\theta, k}\) represents the velocity along the reference path and \(\Delta t\) indicates the time interval between two time steps. After obtaining the distance from the reference path and the risk potential field at each future time step, the final MPCC problem is:
	\begin{align}
		\text{min}\hspace{0.5em} & \sum_{k=0}^{T}\left \|\hat{e}_{k}^{c}(\mathbf{x}_{k}) \right \|_{q_{c} }^{2}+\left \|\hat{e}_{k}^{l}(\mathbf{x}_{k})\right \|_{q_{l} }^{2}+\left \|\mathbf{u}_{k}\right \|_{q_{u} }^{2} +q_{r}\gamma ^{k}\Theta (\mathbf{x}_{k}) ^{t} \notag \\
		& -q_{v}v_{\theta, k} \label{eq:objective} \\
		\text{s.t.}\hspace{0.5em} 
		& \mathbf{x}_{0}= \mathbf{x} \label{eq:constraint1} \\
		& \mathbf{x}_{t+1}=f(\mathbf{x}_{t},\mathbf{u}_{t})\label{eq:constraint2}\\
		& \theta _{\mu ,k+1} =\theta _{\mu ,k}+ v_{\theta, k }\Delta t \label{eq:constraint3}\\
		& \underline{\mathbf{x}}\le \mathbf{x}_{k}\le \overline{\mathbf{x}}\label{eq:constraint4} \\
		& \underline{\mathbf{u}}\le \mathbf{u}_{k}\le \overline{\mathbf{u}}\label{eq:constraint5} 
	\end{align}
	In the above nonlinear model predictive control problem(NMPC), \(\mathbf{x}_{t}\) is the state of ego car at time step \(k\). In \eqref{eq:objective}, \(\hat{e}_{k}^{c}(\mathbf{x}_{k})\) and \(\hat{e}_{k}^{l}(\mathbf{x}_{k})\) are adjusted by \(q_{c}\) and \(q_{l}\) respectively which limit the deviation of the vehicle position from the reference path. The term \(\left \|\mathbf{u} _{k}\right \|_{q_{u} }^{2}\) is the input cost which represents the penalty for changes in steering angle and longitudinal velocity because we want to get a smoother trajectory. The inclusion of progress penalty term \(v_{\theta, k}\) in the objective function aims to maximize the vehicle's forward progress along the reference path and the risk term \(\gamma ^{k}\Theta (\mathbf{x}_{k}) ^{t}\) prevents the ego vehicle from getting too close to other vehicles. By adjusting parameters, a trade-off between risk and progress can be achieved, ensuring that the vehicle's behavior is not overly conservative. Equation \eqref{eq:constraint2} is the discretization of the nonlinear model \eqref{eq:model}, \eqref{eq:constraint4} and \eqref{eq:constraint5} limit states and inputs not to exceeding the physical limits of the vehicle. 
	
	\section{EXPERIMENTAL RESULTS}
	\subsection{Implementation Details}
	\textbf{Train:} Our trajectory prediction model is deployed on pytorch\cite{paszke2019pytorch} and trained on the nuScenes dataset\cite{caesar2020nuscenes}. This dataset annotates the historical trajectory of the predicted target for 2 seconds and the real trajectory for the next 6 seconds at a frequency of 2 Hz. We use the nuScenes devkit\cite{github} to get the detailed annotation information and draw graphics. The weight coefficient of the loss function is set to \(\alpha =0.5\) and \(\beta = 1.0\). 
	
	\textbf{Metrics:} We use the MissRate\_2\_k and OffRoadRate to evaluate our prediction model. If the maximum pointwise L2 distance between the prediction and ground truth is greater than 2 meters, the prediction is defined as a miss. So the MissRate\_2\_k can reflect whether the prediction results are comprehensive enough to cover the real trajectory. And the OffRoadRate measures the ratio of predicted trajectories off the road.
	
	\textbf{Planning:} The MPCC motion planner runs with a horizon length of \(N=80\) with a sampling time of \(T_{s}=50ms\), which corresponds to a 4s ahead prediction. The overall framework of our code is implemented using python, while the motion planner is developed in C++. So we use pybind\cite{pybind} to pass the python object to the motion planning function. 
	\subsection{Simulation Platform and Environment}
	The experiments are conducted on the Carla\cite{dosovitskiy2017carla} simulation environment. The simulation world is set to synchronous mode and the elapsed time between two simulation steps is fixed at 0.05s. We set the operating mode of other vehicles to autopilot\footnote{\url{https://carla.readthedocs.io/en/0.9.15/adv_traffic_manager}} and they have a 50\% chance of ignoring the vehicles in front of them, which makes the vehicles' intentions uncertain. Our experiments are conducted on a laptop configured with Intel Core i9-13900HX CPU and 32GB RAM.
	\begin{figure}[tb]
		\centering
		\includegraphics[width=\linewidth]{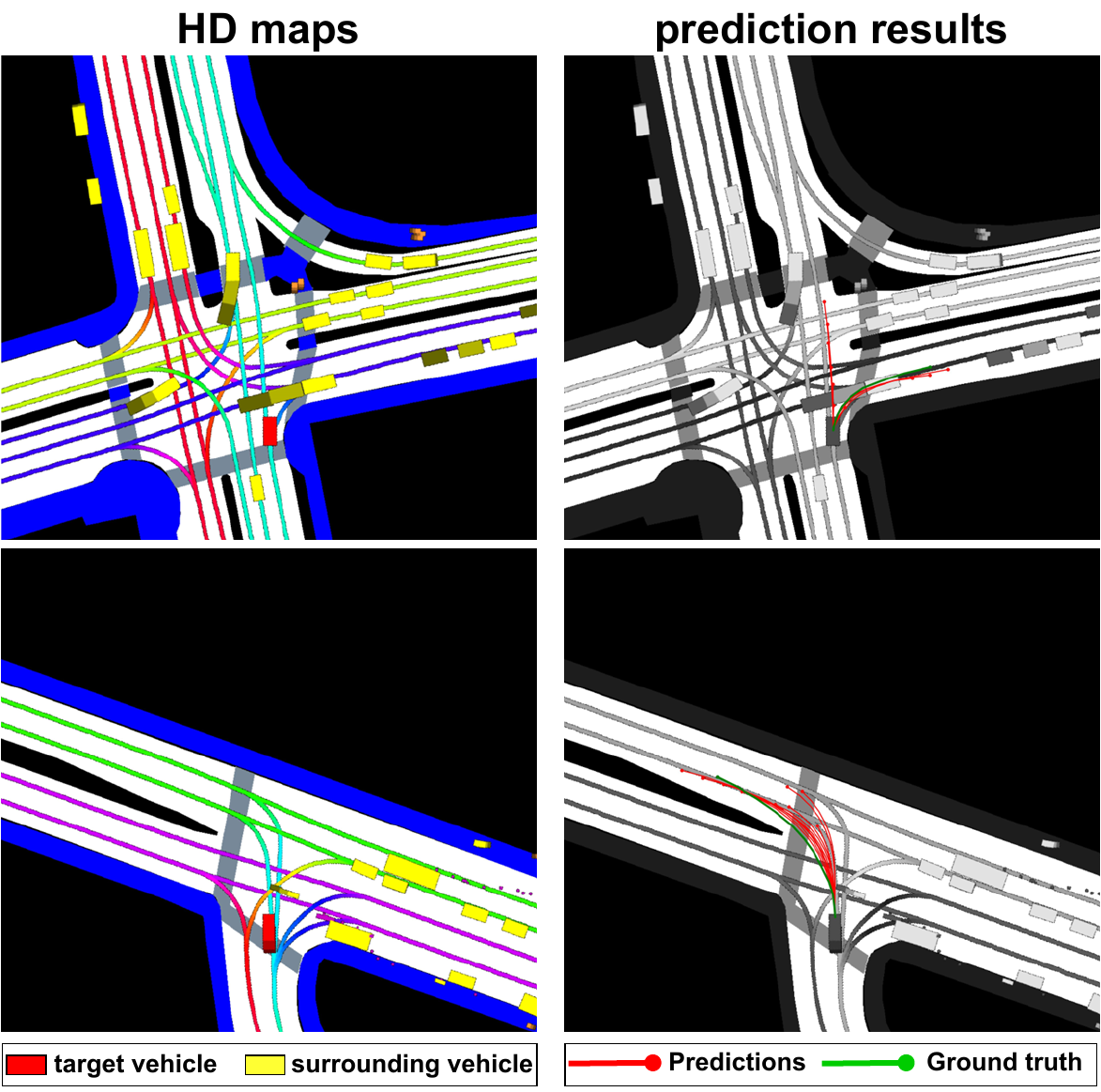}
		\caption{Qualitative results of our proposed prediction model. The pictures in the left column are HD maps of nuScenes in different scenes, The red box in a HD map represents the predicted target and the yellow boxes represent other vehicles. Different lanes are marked with lines of different colors, with white region indicating drivable areas and blue region indicating pedestrian zones. The pictures in the right column show the top 10 predicted trajectories with the highest probability and ground truth in different scenes.}
		\label{fig:pre_result}
	\end{figure}
	\subsection{Prediction Results}
	We report the comparison results with other advanced models on the nuScenes leaderboard in \cref{tab:nuScenes comparision}. The benchmark is ranked by MR\_2\_5, where we rank first. And we achieve comparable results in MR\_2\_10 and OffRoadRate. This proves that our prediction model can obtain relatively comprehensive trajectory predictions. 
	\begin{table}[tb]
		\centering
		\setlength{\fboxrule}{1.5pt}
		\setlength{\fboxsep}{1pt}
		\captionsetup{singlelinecheck=false, justification=centering}
		\caption{Comparison with models on nuScenes}
		\label{tab:nuScenes comparision}
		\begin{tabular}{cccc}
			\hline
			\textbf{Methods} & \textbf{MR\_2\_5(\%)} & \textbf{MR\_2\_10(\%)} & \textbf{OffRoadRate} \\
			\hline
			\textbf{TRTP} & \fcolorbox{red}{white}{\textbf{45.61}} & \fcolorbox{red}{white}{32.81} & \fcolorbox{red}{white}{0.015}\\
			DGCN\_ST\_LANE & 46.99 & 42.22 & 0.124\\
			DSS & 49.23 & \textbf{30.99} & 0.018 \\
			PGP & 51.90 & 34.34 & 0.028 \\
			LaPred++ & 52.60 & 46.19 & 0.060\\
			LaPred & 52.60 & 46.19 & 0.091\\
			CASPNet\_v2 & 53.15 &31.81 & \textbf{0.011}\\
			\hline
		\end{tabular}
	\end{table}
	
	We present the qualitative prediction results of our proposed model in \autoref{fig:pre_result}, where the top 10 trajectories with the highest probability are plotted. This shows that our prediction model can obtain multi-modal predicted trajectories according to different target regions, and some trajectories are close enough to the ground truth. 
	\subsection{Qualitative Results}
	In order to verify the feasibility of our method, we conduct qualitative experiments in an unprotected left turn scenario and a merging scenario.
	\subsubsection{Unprotected left turn}
\begin{figure}[b]
		\centering
		\begin{subfigure}{0.24\linewidth}
			\centering
			\includegraphics[width=\textwidth]{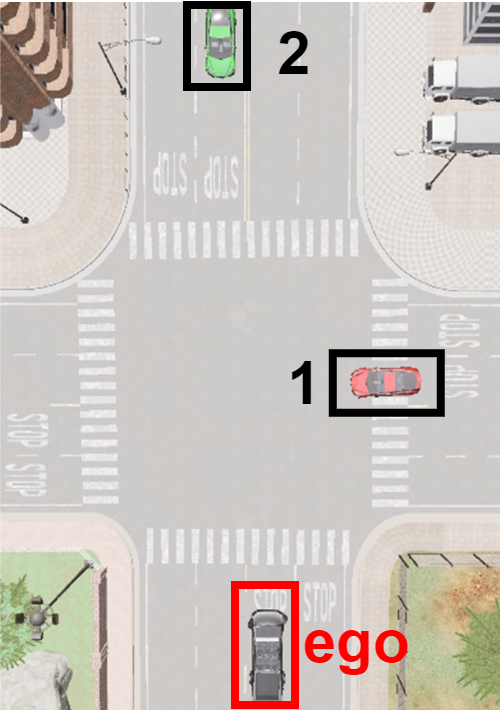}
			\caption{}
			\label{fig:0a}
		\end{subfigure}
		\begin{subfigure}{0.24\linewidth}
			\centering
			\includegraphics[width=\textwidth]{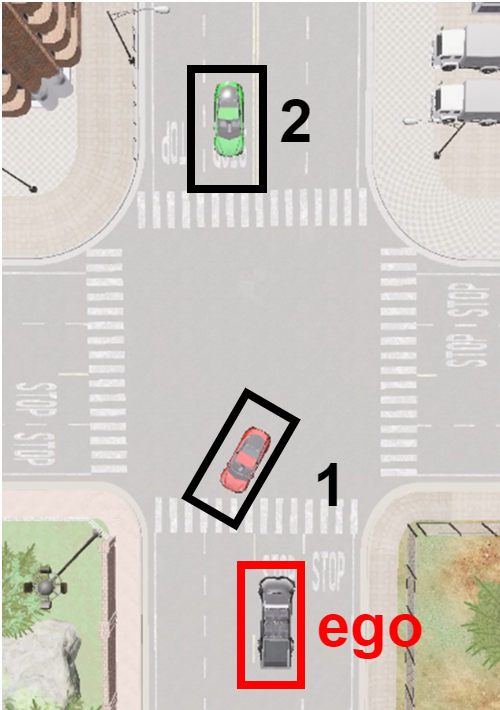}
			\caption{}
			\label{fig:0b}
		\end{subfigure}
		\begin{subfigure}{0.24\linewidth}
			\centering
			\includegraphics[width=\textwidth]{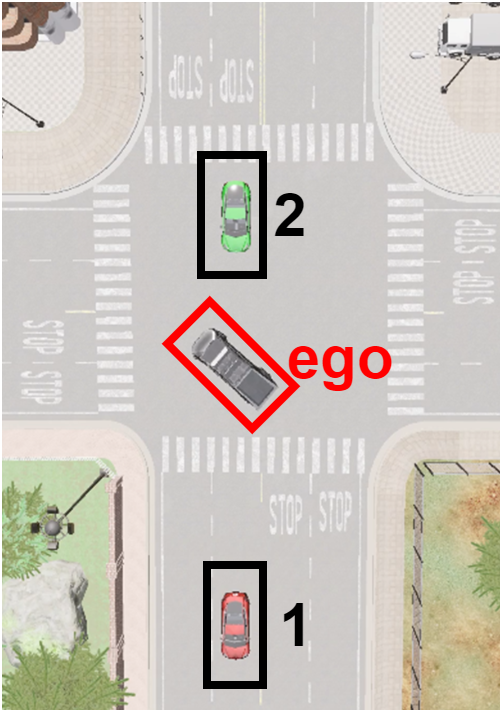}
			\caption{}
			\label{fig:0c}
		\end{subfigure}
		\begin{subfigure}{0.24\linewidth}
			\centering
			\includegraphics[width=\textwidth]{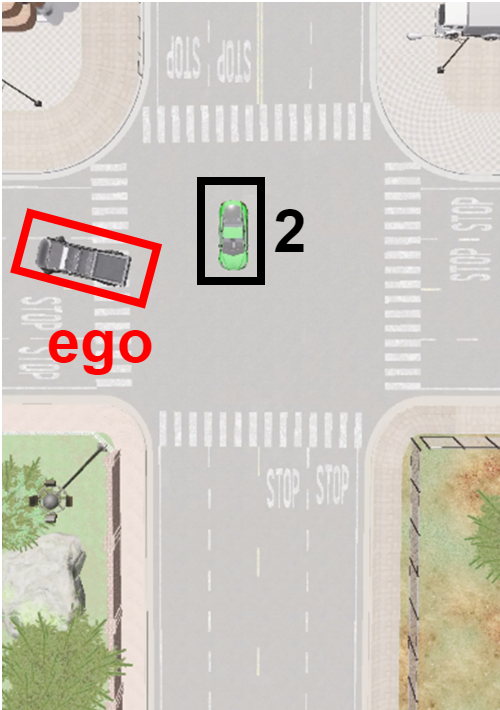}
			\caption{}
			\label{fig:0d}
		\end{subfigure}
		\caption{Key frames of an unprotected left turn scenario. The ego vehicle is marked with a red rectangle. The vehicle 1 and vehicle 2 that will interact with the ego vehicle are marked with black rectangles.}
		\label{fig:unprotected left turn}
	\end{figure}

	As shown in \autoref{fig:unprotected left turn}, the ego vehicle enters an intersection without traffic lights. In order to better reflect the danger of the scenario, we set the probability that other vehicles ignore the stop signs to 100\%. There are three vehicles entered the intersection at the same time(a). Firstly, due to the faster turning speed of the vehicle1, the ego vehicle yields the right of way(b). Then the ego vehicle accelerates to make vehicle 2 give way(c), and finally leaves the intersection(d). Experimental results show that our method can avoid danger in time and does not behave conservatively.
	\subsubsection{Merging scenario}
\begin{figure*}[tb]
		\centering
		\begin{subfigure}{0.32\textwidth}
			\centering
			\includegraphics[width=\textwidth]{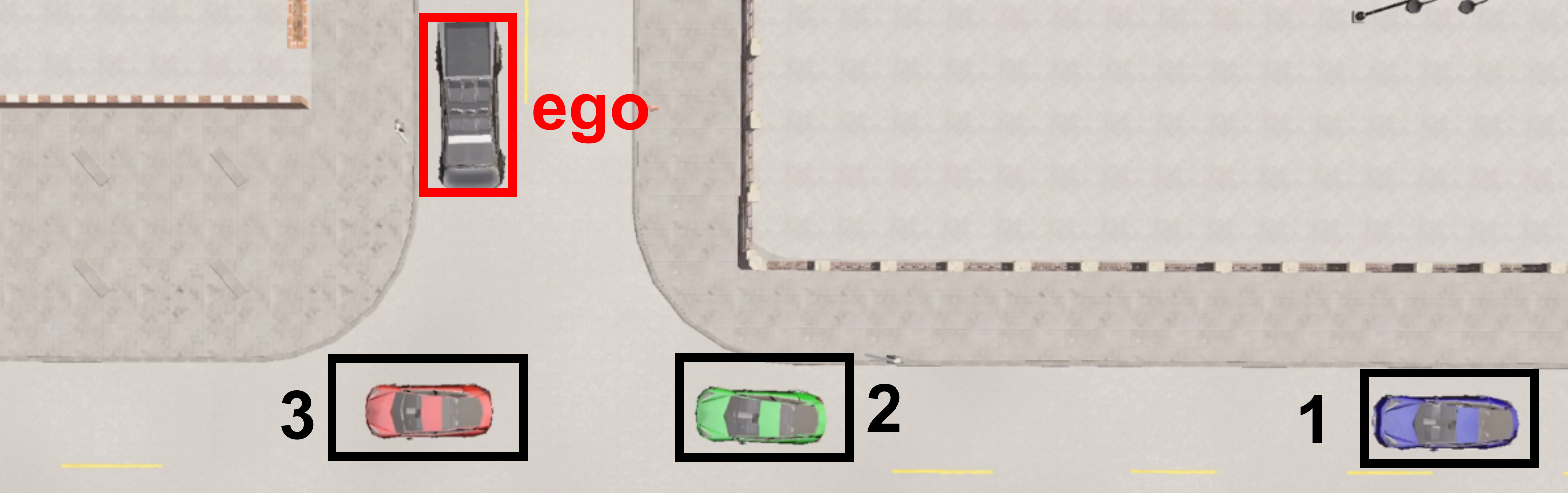}
			\caption{}
			\label{fig:1a}
		\end{subfigure}%
		\hfill
		\begin{subfigure}{0.32\textwidth}
			\centering
			\includegraphics[width=\textwidth]{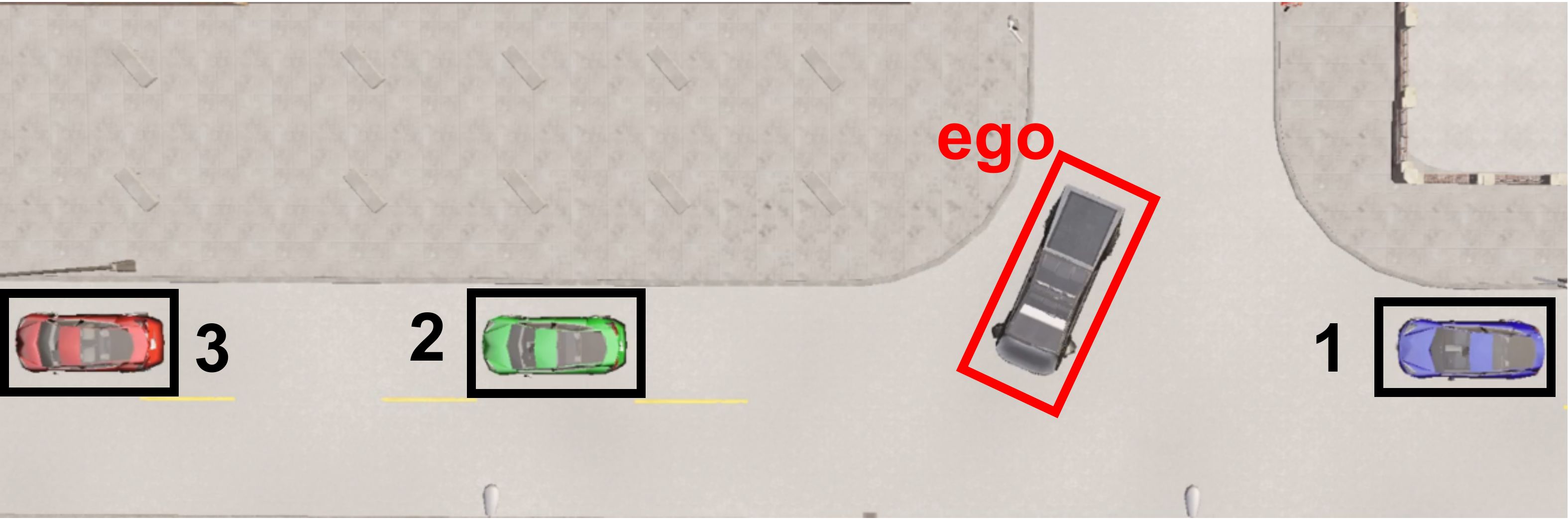}
			\caption{}
			\label{fig:1b}
		\end{subfigure}%
		\hfill
		\begin{subfigure}{0.32\textwidth}
			\centering
			\includegraphics[width=\textwidth]{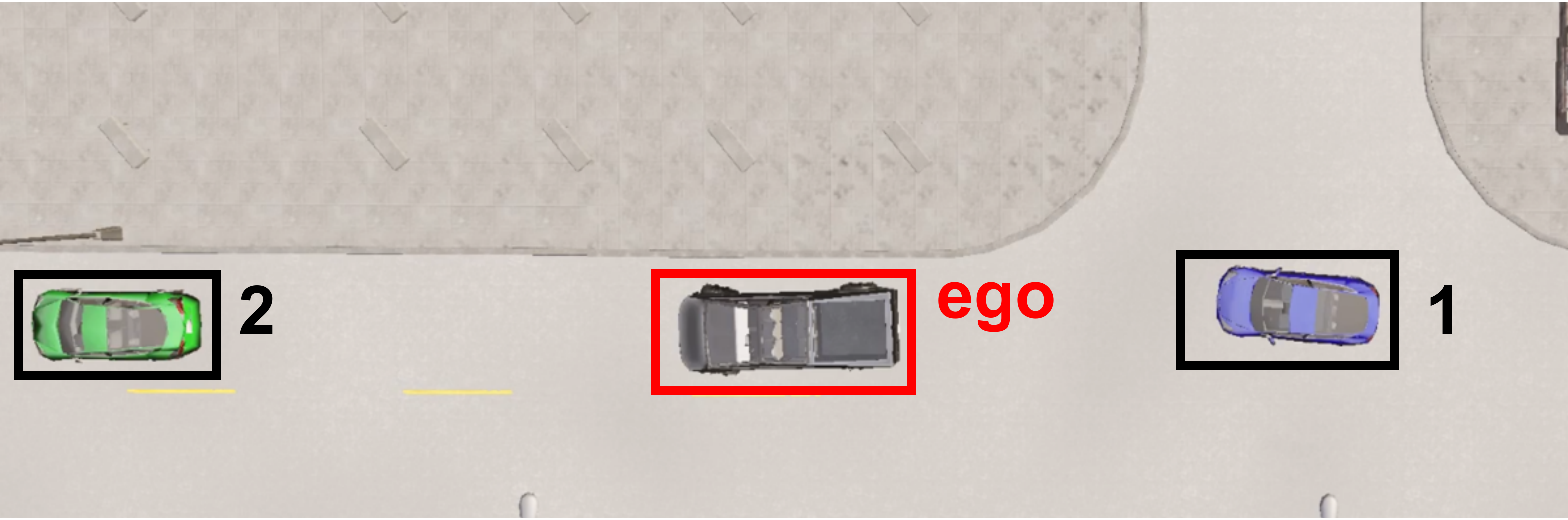}
			\caption{}
			\label{fig:1c}
		\end{subfigure}%
		\caption{Key frames of a merging scenario. The ego vehicle is marked with a red rectangle and the other three vehicles in the traffic flow are all marked with black rectangles.}
		\label{fig:merging}
	\end{figure*}

	As shown in \autoref{fig:merging}, the ego vehicle enters a T-intersection and merges safely into the gap between two other vehicles. We set the probability of other vehicles ignoring traffic lights to 100\% to reflect the authenticity of the traffic situation and increase the risk. When the scenario is initialized, the car is parked 11m away from the T-intersection and decides when to merge into the traffic flow. Firstly, due to the aggressive behavior of vehicle 2, the ego vehicle waits for it to pass by(\ref{fig:1a}). After the vehicle 2 overtakes, the ego vehicle decides to merge(\ref{fig:1b}), and finally, the ego vehicle safely completed its task of merging into the traffic flow(\ref{fig:1c}). The results show that our method can safely and efficiently plan reasonable trajectories when facing complex interaction scenarios.
	\subsection{Quantitative Results}
	To quantitatively evaluate the performance of the proposed method we conduct comparative experiments with GPIR\cite{cheng2022real} and ablation experiments on Carla.
	\subsubsection{Comparison experiments}
	We conduct comparative experiments in an unprotected left turn and a merging scenario mentioned above. The ego vehicle navigates the scenario following the global route and repeats 100 times. To validate the performance of our method, we choose average speed(Avg Spd) and total scenario time(Time) to measure efficiency and collision rate(Col Rate) to measure security. 
	As shown in \cref{tab:Comparision experiments}, our method outperforms GPIR in Avg Spd and Time, and reaches 0\% Col Rate in both scenarios. This indicates that our approach can ensure driving safety while maintaining driving efficiency. In the two different scenarios, GPIR also achieved a collision rate of 0\% because it always waits for all vehicles in front to pass, which leads to overly conservative behavior. 
	\begin{table}[tb]
		\centering
		\captionsetup{singlelinecheck=false, justification=centering}
		\caption{Comparison experiments}
		\label{tab:Comparision experiments}
		\begin{tabular}{c|c|ccc}
			\hline
			\multicolumn{2}{c|}{\textbf{Methods}} & \textbf{\makecell[c]{Time\\(s)}} & \textbf{\makecell[c]{Avg Spd\\(m/s)}} &\textbf{\makecell[c]{Col Rate\\(\%)}}\\
			\hline
			\multirow{2}{*}{\makecell[c]{Unprotected\\left turn}} & GPIR &7.5 & 4.29 & \textbf{0} \\
			& \textbf{Ours} &\textbf{7.0} & \textbf{4.60} & \textbf{0}\\
			\hline
			\multirow{2}{*}{Merging} & GPIR &8.3 & 2.20 & \textbf{0}\\
			& \textbf{Ours} &\textbf{6.8} & \textbf{2.68} & \textbf{0}\\
			\hline
		\end{tabular}
	\end{table}
	
	\subsubsection{Ablation experiments} 
	We conduct ablation experiments using the same metrics as the quantitative experiments in the merging scenario. To verify the advantages of our trajectory prediction model, constant speed prediction(CSP) and TRTP with the same planning module mentioned in \ref{sec:Motion planning} will be experimented in this scenario. To ensure fairness in comparison, the planning module utilizes the same set of parameters. As shown in \cref{tab:Ablation experiments}, compared with the CSP, the TRTP plus MPCC reduces the Col Rate and slightly better in terms of Avg Spd and Time. The experimental results prove the effectiveness of TRTP.
	\begin{table}[tb]
		\centering
		\captionsetup{singlelinecheck=false, justification=centering}
		\caption{Ablation experiments}
		\label{tab:Ablation experiments}
		\begin{tabular}{cccc}
			\hline
			\textbf{Methods} & \textbf{Time(s)} & \textbf{Avg Spd(m/s)} & \textbf{Col Rate(\%)}\\
			\hline
			CSP+MPCC &7.0 &2.64 &5\\
			\textbf{TRTP+MPCC(ours)} & \textbf{6.8} & \textbf{2.68} & \textbf{0}\\
			\hline
		\end{tabular}
	\end{table}
	\section{CONCLUSIONS AND FUTURE WORK}
	In this paper, we present an integrating comprehensive trajectory prediction with risk potential field method for autonomous driving. The approach combines our proposed trajectory prediction model TRTP with risk potential field-based trajectory planning, achieving a balance between safety and efficiency in interactive scenarios. Qualitative and quantitative experiments demonstrate the safety and effectiveness of our method. In the future, we will conduct experiments in more scenarios.
	\bibliographystyle{plain}
	\bibliography{1}
	
\end{document}